\documentclass{article}

\usepackage{times}
\usepackage{graphicx}
\usepackage{booktabs}
\usepackage{subfigure}
\usepackage{natbib}
\newcommand{\citeg}[1]{\citep[e.g.,][]{#1}}
\usepackage{microtype}
\usepackage{amsmath, amsfonts}
\usepackage{color,framed,soul}

\usepackage[accepted]{icml2014}

\usepackage{algorithm}
\usepackage{algorithmicx}
\usepackage{algpseudocode}
\usepackage{placeins}

\algrenewcommand\algorithmicindent{0.9em}%

\newcommand{\TODO}[1]{}
\newcommand{\NOTE}[1]{}
\renewcommand{\TODO}{\textcolor{red}}
\renewcommand{\NOTE}{\textcolor{blue}}
\newcommand{\ie}{i.e.\ }

\newcommand{\realdomain}{\mathbb{R}}
\newcommand{\given}{\,|\,}

\renewcommand{\vec}[1]{\boldsymbol{#1}}
\newcommand{\mat}[1]{\boldsymbol{#1}}

\newcommand{\sigm}{\operatorname{sigm}}

\newcommand{\loss}{\mathcal{J}_{\scriptscriptstyle OA}}

\newcommand{\expectation}{\mathop{\mathbb{E}}\limits}
\newcommand{\params}{\vec{\theta}}

\newcommand{\xogtd}{\vec{x}_{o_{>d}}}
\newcommand{\xoltd}{\vec{x}_{o_{<d}}}
\newcommand{\xod}{x_{o_d}}

\icmltitlerunning{A Deep and Tractable Density Estimator}

\begin{document}

\twocolumn[
\icmltitle{A Deep and Tractable Density Estimator}

\icmlauthor{Benigno Uria}{b.uria@ed.ac.uk}
\icmlauthor{Iain Murray}{i.murray@ed.ac.uk}
\icmladdress{School of Informatics, University of Edinburgh}
\icmlauthor{Hugo Larochelle}{hugo.larochelle@usherbrooke.ca}
\icmladdress{D{\'e}partement d'informatique, Universit{\'e} de Sherbrooke}

\icmlkeywords{Deep Learning, Neural Networks, Density Estimation}

\vskip 0.3in
]

\begin{abstract}
The Neural Autoregressive Distribution Estimator (NADE) and its
real-valued version RNADE are competitive density models of
multidimensional data across a variety of domains. These models use a
fixed, arbitrary ordering of the data dimensions. One can easily
condition on variables at the beginning of the ordering, and
marginalize out variables at the end of the ordering, however other
inference tasks require approximate inference. In this work we
introduce an efficient procedure to simultaneously train a NADE model
for each possible ordering of the variables, by sharing parameters
across all these models. We can thus use the most convenient model for
each inference task at hand, and ensembles of such models with
different orderings are immediately available. Moreover, unlike the
original NADE, our training procedure scales to deep models.
Empirically, ensembles of Deep NADE models obtain state of the art
density estimation performance.
\end{abstract}

\section{Introduction}

In probabilistic approaches to machine learning, large collections of variables
are described by a joint probability distribution. There is considerable
interest in flexible model distributions that can fit and generalize from
training data in a variety of applications. To draw inferences from these
models, we often condition on a subset of observed variables, and report the
probabilities of settings of another subset of variables, marginalizing out any
unobserved nuisance variables. The solutions to these inference tasks often
cannot be computed exactly, and require iterative approximations such as Monte
Carlo or variational methods \citeg{bishop2006}. Models for which inference is
tractable would be preferable.

NADE~\cite{Larochelle+Murray-2011}, and its real-valued variant RNADE~\cite{UriaB2013}, have been shown to be
state of the art joint density models for a variety of real-world datasets, as
measured by their predictive likelihood.
These models predict each variable sequentially in an arbitrary order, fixed at
training time. Variables at the beginning of the order can be set to observed
values, i.e., conditioned on. Variables at the end of the ordering are not
required to make predictions; marginalizing these variables requires simply
ignoring them. However, marginalizing over and conditioning on any arbitrary
subsets of variables will not be easy in general.

In this work, we present a procedure for training a factorial number of
NADE models simultaneously; one for each possible ordering of the variables.
The parameters of these models are shared, and we optimize the mean cost over all
orderings using a stochastic gradient technique.
After fitting the shared parameters, we can
extract, in constant time, the NADE model with the variable ordering that
is most convenient for any given inference task. While the different NADE models
might not be consistent in their probability estimates, this property is
actually something we can leverage to our advantage, by generating ensembles of
NADE models ``on the fly'' (i.e., without explicitly training any such ensemble)
which are even better estimators than any single NADE\@. In addition, our
procedure is able to train a deep version of NADE incurring an extra
computational expense only linear in the number of layers.

\section{Background: NADE and RNADE}

Autoregressive methods use the product rule to factorize the probability density
function of a $D$-dimensional vector-valued random variable $\vec{x}$ as a
product of one-dimensional conditional distributions:
\begin{equation}
p(\vec{x}) = \prod_{d=1}^{D}p(\xod \given \xoltd)\mathrm{,}\label{autoregressive-model}
\end{equation}
where $o$ is a $D$-tuple in the set of
permutations of $(1,\ldots,D)$ that serves as an ordering of the elements in
$\vec{x}$, $\xod$ denotes the element of $\vec{x}$ indexed by the $d$-th
 element  in $o$, and $\xoltd$ the elements of $\vec{x}$ indexed by the first
$d-1$ elements in $o$. This factorisation of the pdf assumes no conditional
independences. The only element constraining the modelling ability of an
autoregressive model is the family of distributions chosen for each of the
conditionals.

In the case of binary data, autoregressive models based on logistic regressors
and neural networks have been proposed~\cite{Frey98,bengio:2000:nips}. The neural autoregressive density
estimator~(NADE)~\cite{Larochelle+Murray-2011}, inspired by a mean-field approximation to the
conditionals of Equation~\eqref{autoregressive-model} of a restricted Boltzmann machine
(RBM), uses a set of one-hidden-layer neural networks with tied parameters to
calculate each conditional:
\begin{align}
p(\xod = 1 \given
\xoltd) & = \sigm(\mat{V}_{\cdot,o_d}\vec{h}_d+b_{o_d})\label{eq:nade-output}\\
\vec{h}_d & = \sigm(\mat{W}_{\cdot, o_{<d}}\xoltd + \vec{c}),
\label{eq:hidden-units}
\end{align}
where $H$ is the number of hidden units, and $\mat{V} \in \realdomain^{H\times
D}$, $\vec{b} \in \realdomain^D$, $\mat{W} \in \realdomain^{H \times D }$,
$\vec{c} \in \realdomain^H$ are the parameters of the NADE model.

A NADE can be trained by regularized gradient descent on the negative
log-likelihood given the training dataset $\vec{X}$.%

In NADE the activation of the hidden units in \eqref{eq:hidden-units} can be
computed recursively:
\begin{align}
\vec{h}_d & = \sigm(\vec{a}_{d}) \qquad \text{where} \qquad
\vec{a}_1 = \vec{c} \label{eq:nade-recursive-init} \\
\vec{a}_{d+1} & = \vec{a}_d+\xod\vec{W}_{\cdot,o_d}.
\label{eq:tied-activations}
\end{align}
This relationship between activations allows faster training and evaluation of a
NADE model, $O(DH)$, than autoregressive models based on untied neural networks,
$O(D^2H)$.

NADE has recently been extended to allow density estimation of real-valued
vectors~\cite{UriaB2013} by using mixture density networks or MDNs~\cite{BishopC1994} for each of the
conditionals in Equation~\eqref{autoregressive-model}. The networks' hidden
layers use the same parameter sharing as before, with activations computed as
in~\eqref{eq:tied-activations}.

NADE and RNADE have been shown to offer better modelling performance than
mixture models and untied neural networks  in a range of datasets.
Compared to binary RBMs with hundreds of hidden units, NADEs usually have
slightly worse modelling performance, but they have three desirable
properties that the former lack: 1)~an easy training procedure by gradient
descent on the negative likelihood of a training dataset, 2)~a tractable
expression for the density of a datapoint, 3)~a direct ancestral sampling
procedure, rather than requiring Markov chain Monte Carlo methods.

Inference under a NADE is easy as long as the variables to condition on are at
the beginning of its ordering, and the ones to marginalise over are at the end.
To infer the density of $\vec{x}_{o_a\ldots o_b}$ while
conditioning on $\vec{x}_{o_1\ldots o_{a-1}}$, and marginalising over
$\vec{x}_{o_{b+1\ldots D}}$, we simply write
\begin{align}
    p(\vec{x}_{o_{a\ldots b}} \given \vec{x}_{o_{1\ldots a-1}}) &=
    \prod_{d=a}^b p(\xod \given \xoltd)\mathrm{,} \label{eq:easy-marginalization}
\end{align}
where each one-dimensional conditional is directly available from the model.
However, as in most models, arbitrary probabilistic queries require approximate
inference methods.

A disadvantage of NADE compared to other neural network models is that an efficient
deep formulation \citeg{bengio2009learning} is not available. While extending NADE's definition to multiple hidden layers is trivial
(we simply introduce regular feed-forward layers between the computation
of Equation~\ref{eq:hidden-units} and of Equation~\ref{eq:nade-output}),
we lack a recursive expression like Equations~\ref{eq:nade-recursive-init}~and~\ref{eq:tied-activations}
for the added layers. Thus, when NADE has more than
one hidden layer, each additional hidden layer must be computed separately for each input dimension,
yielding a complexity cubic on the
size of the layers $O(DH^2L)$, where $L$ represents the number of layers. This
scaling seemingly made a deep NADE impractical, except for datasets of low dimensionality.

\section{Training a factorial number of NADEs}

Looking at the simplicity of inference in
Equation~\eqref{eq:easy-marginalization}, a naive approach that could exploit
this property for any inference task would be to train as many NADE
models as there are possible orderings of the input variables. Obviously,
this approach, requiring $O(D!)$ time and memory, is not viable. However, we
show here that through some careful parameter tying between models, we can
derive an efficient stochastic procedure for training all models, minimizing
the mean of their negative log-likelihood objectives.

Consider for now a parameter tying strategy that simply uses the same
weight matrices and bias parameters across all NADE models (we will
refine this proposal later). We will now write $p(\vec{x} \given
\params, o)$ as the joint distribution of the NADE model that uses
ordering $o$ and $p(x_{o_d}^{(n)} \given \vec{x}_{o_{<d}}^{(n)},
\params, o_{<d}, o_d )$ as its associated conditionals, which are
computed as specified in
Equations~\eqref{eq:nade-output}~and~\eqref{eq:hidden-units}, or their
straightforward extension in the deep network case. Thus we explicitly treat
the ordering $o$ as a random variable. Notice that the $d^{\rm th}$ conditional
only depends on the first $d$ elements of the ordering, and is thus
exactly the same across NADE models sharing their first $d$ elements in $o$.
 During training we will
attempt to minimise the expected (over variable orderings) negative
log-likelihood of the model for the training data:
\newcommand{\expectationorderings}{\expectation_{o \in D!}}
\newcommand{\expectationoltd}{\expectation_{o_{<d}}}
\newcommand{\expectationogtd}{\expectation_{o_{>d}}}
\newcommand{\expectationod}{\expectation_{o_{d}}}
\newcommand{\expectationtrainingdata}{\expectation_{\vec{x^{(n)}}\in\vec{X}}}
\newcommand{\sumtrainingdata}{\sum_{\vec{x^{(n)}}\in\vec{X}}}
\begin{align}
\loss(\params) & = \expectationorderings -\log p(\vec{X}
\given \params, o) \\
& \propto
\expectationorderings \expectationtrainingdata -\log
p(\vec{x^{(n)}} \given \params, o), \label{eq:loss-function}
\end{align}
where $D!$ is the set of all orderings
(\ie permutations of $D$ elements).
This objective does not correspond to a mixture model, in which case the
expectation over orderings would be inside the $\log$ operation.

Using NADE's autoregressive expression for the density of a datapoint,
\eqref{eq:loss-function} can be rewritten as:
\begin{equation}
\loss(\params)=\expectationorderings \expectationtrainingdata
\sum_{d=1}^D -\log p(x_{o_d}^{(n)} \given \vec{x}_{o_{<d}}^{(n)}, \params, o).
\end{equation}
Where $d$ indexes the elements in the order, $o$, of the dimensions. By moving
the expectation over orders inside the sum over the elements of the order,
the order can be split in three parts:
$o_{<d}$ standing for the index of the $d\!-\!1$ first dimensions in the ordering; $o_d$ the
index of the $d$-th dimension in the ordering, and $o_{>d}$ standing for the
indices of the remaining dimensions in the ordering. Therefore, the loss function
can be rewritten as:
\begin{equation}\begin{split}
\hspace{-1ex}\loss(\params)= \hspace{-2ex}\expectationtrainingdata\sum_{d=1}^D
\expectationoltd \expectationod \expectationogtd - \log
p(x_{o_d}^{(n)} \given \vec{x}_{o_{<d}}^{(n)}, \params, o_{<d}, o_d )
\end{split}\end{equation}
the value of each term does not depend on $o_{>d}$. Therefore, it can be
simplified as:
\begin{equation}
\loss(\params)= \expectationtrainingdata\sum_{d=1}^D \expectationoltd
\expectationod - \log p(x_{o_d}^{(n)} \given \vec{x}_{o_{<d}}^{(n)}, \params, o_{<d}, o_d )
\label{eq:loss-intractable}
\end{equation}
In practice, this loss function \eqref{eq:loss-intractable} will have a very
high number of terms and will have to be approximated by sampling $\vec{x}^{(n)}$, $d$, and $o_{<d}$.
The innermost expectation over values of $o_d$ can be calculated cheaply
for a NADE given that the hidden unit states~$\vec{h}_d$ are shared for all
possible $o_d$.
Therefore, assuming all orderings are equally probable, we will estimate $\loss(\params)$ by:
\begin{align}
\widehat{\loss}(\params) &= \dfrac{D}{D-d+1}\sum_{o_d} -\log
p(x_{o_d}^{(n)} \given \xoltd^{(n)}, \params, o_{<d}, o_d )
\label{eq:hatloss}
\end{align}
 which provides an unbiased estimator of \eqref{eq:loss-function}.
Thus training can be done by descent on the stochastic gradient of
$\widehat{\loss}(\params)$.
An implementation of this order-agnostic training procedure corresponds to an
artificial neural network with $D$ inputs and $D$ outputs (or an MDN in the
real-valued case), where the input values in $o_{\ge d}$ have been set to zero
and gradients are backpropagated only from the outputs in $o_{\ge d}$, and
rescaled by $\tfrac{D}{D-d+1}$.

The end result is a stochastic training update costing $O(DH +
H^2L)$, as in regular multilayer neural networks.  At test time, we
unfortunately cannot avoid a complexity of $O(DH^2L)$ and perform $D$
passes through the neural network to obtain all $D$ conditionals for
some given ordering. However, this is still tractable, unlike
computing probabilities in a restricted Boltzmann machine or
a deep belief network.

\subsection{Improved parameter sharing using input masks}

While the parameter tying proposed so far is simple, in practice it leads to
poor performance. One issue is that the values of the hidden units, computed
using \eqref{eq:hidden-units}, are the same when a dimension is in $\xogtd$ (a
value to be predicted) and when the value of that dimension is zero
and conditioned on. When
training just one NADE with a fixed $o$, each output unit knows which inputs
feed into it, but in the multiple ordering case that information is lost when
the input is zero.

\newcommand{\mask}{m_{o_{<d}}}
In order to make this distinction possible, we augment the parameter sharing
scheme by appending to the inputs a binary mask vector $\mask \in \{0,1\}^D$
indicating which dimensions are present in the input. That is, the $i$-th
element of $\mask$ is $1$ if $i \in o_{<d}$ and $0$ otherwise.
One interpretation of this scheme is that the bias vector $\vec{c}$ of the first
hidden layer is now dependent on the ordering $o$ and the value of $d$, thus
slightly relaxing the strength of parameter sharing between the NADE models.
We've found in practice that this adjustment is crucial to obtain good
estimation performance. Some results showing the difference in statistical
performance with and without training masks can be seen in
Table~\ref{tab:MNIST-results} as part of our experimental analysis (see
Section~\ref{sec:experiments} for details).

\section{On the fly generation of NADE ensembles}
Our order-agnostic training procedure can be thought of as
producing a set of parameters that can be used by a factorial number
of NADEs, one per ordering of the input variables.  These different
NADEs will not, in general, agree on the probability of a given
datapoint. While this disagreement might look unappealing at first, we can
actually use this source of variability to our advantage, and obtain better
estimates than possible with a set of consistent models.

A NADE with a given input ordering corresponds to a different hypothesis
space than other NADEs with different ordering. In other words, each
NADE with a different ordering is a model in its own right, with slightly different inductive bias, despite
the parameter sharing.

A reliable approach to improve on some given estimator
is to instead construct an ensemble of multiple, strong but different
estimators, e.g.\ with bagging~\cite{Ormoneit1995} or stacking~\cite{Smyth1999}.
Our training procedure suggest a straightforward way of generating
ensembles of NADE models: generate a set of uniformly distributed
orderings $\{o^{(k)}\}_{k=1}^K$ over the input variables and use the
average probability $\frac{1}{K} \sum_{k=1}^K p(\vec{x} | \theta,
o^{(k)})$ as our estimator.

Ensemble averaging increases the computational cost of density estimation linearly
with the size of the ensemble, while the complexity of sampling
doesn't change (we pick an ordering $o^{(k)}$ at random from the
ensemble and sample from the corresponding NADE)\@.  Importantly, the
computational cost of training remains the same, unlike ensemble
methods such as bagging. Moreover, an adequate number of components
can be chosen after training, and can even be adapted to the available
computational budget on the fly.

\section{Related work}

As mentioned previously, autoregressive density/distribution
estimation has been explored before by others. For the binary data case,
\citet{Frey98} considered the use of logistic regression conditional models,
while \citet{bengio:2000:nips} proposed a single layer neural network architecture,
with a parameter sharing scheme different from the one in the
NADE model \citep{Larochelle+Murray-2011}. In all these cases however, a single (usually random)
input ordering was chosen and maintained during training.

\citet{GregorK2011} proposed training a variant of the NADE architecture under
stochastically generated random orderings.
Like us, they observed much worse performance than when choosing a single
variable ordering, which motivates our proposed parameter sharing scheme relying
on input masks.
\citeauthor{GregorK2011} generated a single ordering for
each training update, and conditioned on contexts of all possible sizes to compute
the log-probability of an example and its gradients.  Our stochastic approach
uses only a single conditioning configuration for each update,
but computes the average log-probability for the next dimension under all
possible future orderings. This change allowed us to generalize
NADE to deep architectures with an acceptable computational cost.

\citet{Goodfellow2013} introduced a procedure to train deep Boltzmann
machines by maximizing a variational approximation of their generalised pseudo
likelihood. This results in a training procedure similar to the one presented in
this work, where a subset of the dimension is predicted given the value of the
rest.

Our algorithm also bears similarity with denoising
autoencoders~\cite{VincentPLarochelleH2008} trained using so-called
``masking noise''. There are two crucial differences however. The
first is that our procedure corresponds to training on the
average reconstruction of only the inputs that are missing from the input
layer. The second is that, unlike denoising autoencoders, the NADE
models that we train can be used as tractable density estimators.

\section{Experimental results}
\label{sec:experiments}
\begin{table*}
\caption{Average test-set log-likelihood per datapoint (in nats) of different
models on eight binary datasets from the UCI repository. Baseline results were
taken from \citet{Larochelle+Murray-2011}.}
\label{tab:NADE-UCI-results}
\centering
\begin{tabular}{lcccccccc}
\toprule
{Model} & {Adult} & {Connect4} & {DNA} & {Mushrooms} & {NIPS-0-12} &
{Ocr-letters} & {RCV1} & {Web}\\
\midrule
\begin{small}MoBernoullis\end{small} & $-20.44$ & $-23.41$ & $-98.19$ &
$-14.46$ & $-290.02$ & $-40.56$ & $-47.59$ & $-30.16$ \\
\begin{small}RBM \end{small} & $-16.26$ & $-22.66$ & $-96.74$ &
$-15.15$ & $-277.37$ & $-43.05$ & $-48.88$ & $-29.38$ \\
\begin{small}FVSBN \end{small} & $\mathbf{-13.17}$ & $-12.39$ &
$\mathbf{-83.64}$ & $-10.27$ & $-276.88$ & $-39.30$ & $-49.84$ & $-29.35$ \\
\begin{small}NADE (fixed order) \end{small} & $-13.19$ & $\mathbf{-11.99}$ &
$-84.81$ & $\mathbf{-9.81}$ & $\mathbf{-273.08}$ & $\mathbf{-27.22}$ & $\mathbf{-46.66}$ & $-28.39$ \\
\hline
\begin{small}NADE 1hl\end{small}& $-13.51$ & $-13.04$ & $-84.28$ & $-10.06$ &
$-275.20$ & $-29.05$ & $-46.79 $ & $-28.30$\\
\begin{small}NADE 2hl\end{small}& $-13.53$ & $-12.99$ & $-84.30$ & $-10.05$ &
$-274.69 $ & $-28.92$ & $-46.71$ & $\mathbf{-28.28}$\\
\begin{small}NADE 3hl\end{small}& $-13.54$ & $-13.08$ & $-84.37$ &
$-10.10$ & $-274.86$ & $-28.89$ & $-46.76$ & $-28.29$\\
\begin{small}EoNADE 1hl (2 ord)\end{small}\hspace{-3ex} & $-13.35$ & $-12.81$ &
$-83.52$ & $-9.88$ & $-274.12$ & $-28.36$ & $-46.50$ & $-28.11$
\\
\begin{small}EoNADE 1hl (16 ord)\end{small}\hspace{-3ex} & $-13.19$ & $-12.58$ &
$\mathbf{-82.31}$ & $\mathbf{-9.68}$ & $\mathbf{-272.38}$ & $-27.31$ & $\mathbf{-46.12}$ &
$\mathbf{-27.87}$
\\
\bottomrule
\end{tabular}
\end{table*}
We performed experiments on several binary and real-valued datasets to asses the
performance of NADEs trained using our order-agnostic procedure. We report the
average test log-likelihood of each model, that is, the average log-density of
datapoints in a held-out test set. In the case of NADEs trained in an
order-agnostic way, we need to choose an ordering of the variables so that one
may calculate the density of the test datapoints. We report the average of the
average test log-likelihoods using ten different orderings chosen
at random. Note that this is different from an ensemble, where the
probabilities are averaged before calculating its logarithm.
To reduce clutter,
we have not reported the standard deviation across orderings. In all cases,
this standard deviation has magnitude smaller than the log-likelihood's standard
error due to the finite size of our test sets. These standard errors are also
small enough not to alter the ranking of the different models. In the case of
ensembles of NADEs the standard deviation due to different sets of orderings is,
as expected, even smaller. Every results table is partitioned in two halves, the
top half contains baselines and the bottom half results obtained using our
training procedure. In every table the log-likelihood of the best model, and the
log-likelihood of the best ensemble are shown in bold.%

Training configuration details common to all datasets (except where specified
later on) follow. We trained all order-agnostic NADEs and RNADEs using minibatch
stochastic gradient descent on $\loss$,~\eqref{eq:loss-intractable}. The initial
learning rate, which was chosen independently for each dataset, was reduced
linearly to reach zero after the last iteration. For the purpose of consistency,
we used rectified linear units~\cite{nair2010rectified} in all experiments. We
found that this type of unit allow us to use higher learning rates and made
training converge faster. We used Nesterov's accelerated gradient~\cite{Sutskever2013}
with momentum value 0.9. No weight decay was applied. To avoid
overfitting, we early-stopped training by estimating the log-likelihood on a
validation dataset after each training iteration using the $\widehat{\loss}$
estimator,~\eqref{eq:hatloss}. For models with several hidden layers, each
hidden layer was pretrained using the same hyperparameter values but
only for 20 iterations, see recursive procedure
in Algorithm~\ref{alg:pretraining}.
\begin{algorithm}
\begin{algorithmic}[1]
\caption{Pretraining of a NADE with $n$ hidden layers on dataset X.}
\label{alg:pretraining}
\Procedure{pretrain}{$n$, $X$}
  \If{$n=1$}
    \State\Return{$\Call{random-one-hidden-layer-NADE}$}
  \Else
    \State $nade \gets\Call{pretrain}{n-1}$
    \State $nade \gets\Call{remove-output-layer}{nade}$
    \State $nade \gets\Call{add-a-new-hidden-layer}{nade}$
    \State $nade \gets\Call{add-a-new-output-layer}{nade}$
    \State $nade \gets\Call{train-all}{nade, X, iters=20}$
    \State \Return{$nade$}
  \EndIf
\EndProcedure
\end{algorithmic}
\end{algorithm}

\subsection{Binary datasets}

We start by measuring the statistical performance of a NADE trained using our
order-agnostic procedure on eight binary UCI
datasets~\cite{Bache+Lichman:2013}.%

Experimental configuration details follow. We fixed the number of units per
hidden layer to 500, following \citet{Larochelle+Murray-2011}. We used minibatches
of size 100. Training was run for 100 iterations, each consisting of 1000 weight
updates. The initial learning rate was cross-validated for each of the datasets
among values $\{0.016, 0.004, 0.001, 0.00025, 0.0000675\}$.

Results are shown on Table~\ref{tab:NADE-UCI-results}. We compare our method
to mixtures of multivariate Bernoullis with their number of components
cross-validated among $\{32,64,128,256,512,1024\}$, tractable RBMs of 23 hidden
units, fully visible sigmoidal Bayes networks~(FVSBN), and NADEs trained using a
fixed ordering of the variables. All baseline results are taken
from~\citet{Larochelle+Murray-2011} and details can be found there. NADEs
trained in an order-agnostic manner obtain performances close to those of NADEs
trained on a fixed ordering. The use of several hidden layers offers no
advantage on these datasets. However, ensembles of NADEs obtain higher
test log-likelihoods on all datasets.

We also present results on binarized-MNIST \citep{Salakhutdinov+Murray-2008}, a
binary dataset of 28 by 28 pixel images of handwritten digits. Unlike
classification, density estimation on this dataset remains a challenging task.

Experimental configuration details follow. Training was run for 200 iterations
each consisting of 1000 parameter updates, using minibatches of size 1000. The
initial learning rate was set to $0.001$ and chosen manually by optimizing the
validation-set log-likelihood on preliminary runs.

Results for MNIST are shown in Table~\ref{tab:MNIST-results}. We compare our
method with mixtures of multivariate Bernoulli distributions with 10 and 500
components, fixed-ordering NADEs, RBMs (500 hidden units), and
two-hidden-layer DBNs (500 and 2000 hidden units on each layer) whose
performance was estimated by~\citet{Salakhutdinov+Murray-2008, MurraySal09}. In
order to provide a more direct comparison to our results, we also report the
performance of NADEs trained using a fixed variable-ordering, minibatch
stochastic gradient descent and sigmoid or rectified linear units. We found the
type of hidden-unit did not affect statistical performance, while our minibatch
SGD implementation seems to obtain slightly higher log-likelihoods than
previously reported.

One and two hidden-layer NADEs trained by minimizing $\loss$ obtain marginally
lower (worse) test-likelihoods than a NADE trained for a fixed ordering of the
inputs, but still perform much better than mixtures of multivariate Bernoullis
and very close to the estimated performance of RBMs. More than two hidden layers
are not beneficial on this dataset.

\begin{figure}
\centerline{\includegraphics[width=1.05\columnwidth]{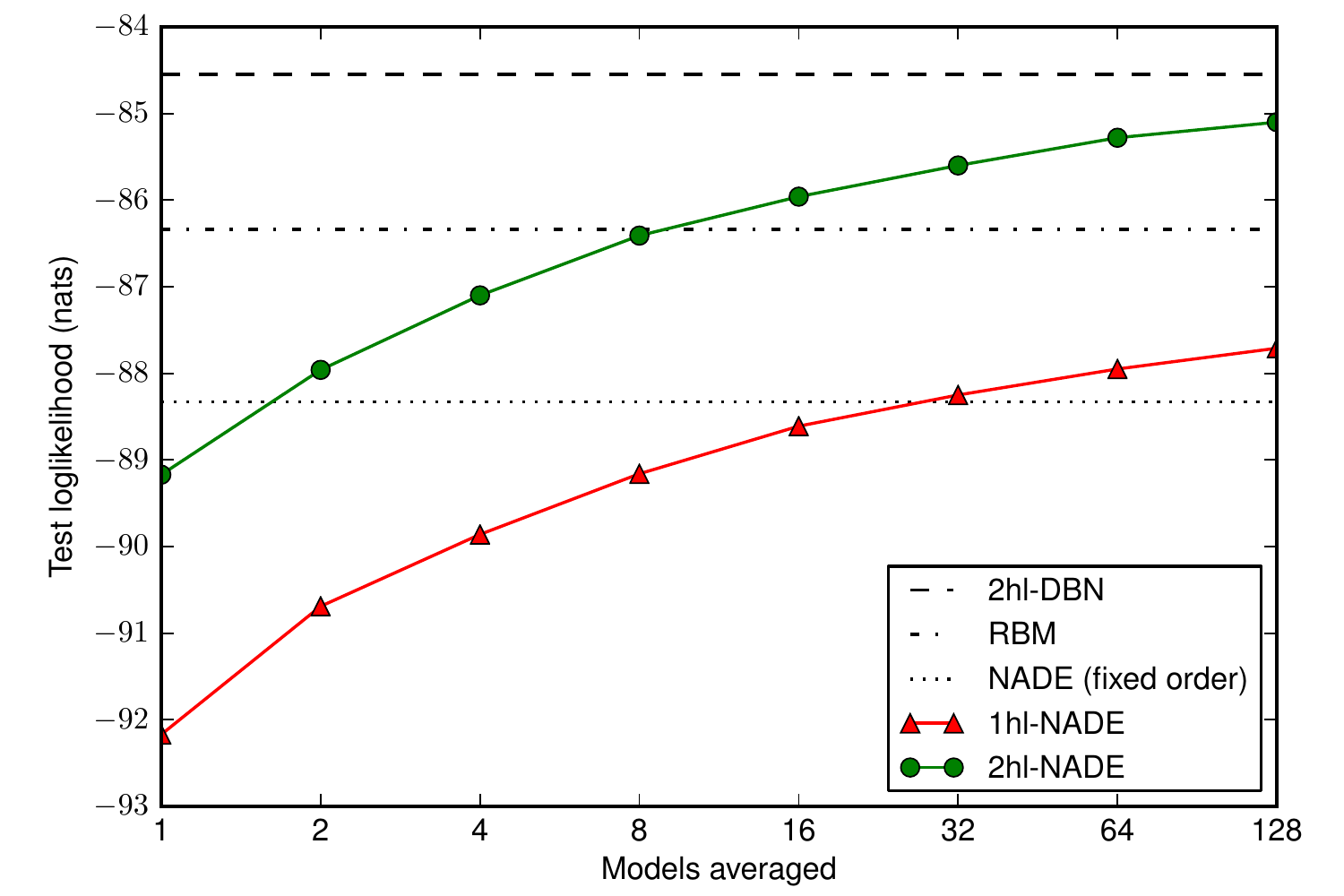}}
\caption{Test-set average log-likelihood per datapoint for RNADEs trained with
our new procedure on binarized images of digits.}
\label{fig:likelihood-vs-orderings}
\end{figure}

Ensembles of NADEs obtained by using NADEs with different variable orderings but
trained simultaneously with our order-agnostic procedure obtain better
statistical performance than NADEs trained using a fixed ordering. These EoNADEs
can also surpass the estimated performance of RBMs with the same number of
hidden units, and even approach the estimated performance of a (larger)
2-hidden-layer deep belief network. A more detailed account of the statistical performance of
EoNADEs can be seen in Figure~\ref{fig:likelihood-vs-orderings}. We also report
the performance on NADE trained by minimizing~$\loss$ but without input masks.
Input masks are necessary for obtaining competitive results.
\begin{table}
\begin{center}
\caption{Average test-set log-likelihood per datapoint of different models
on $28\!\times\!28$ binarized images of digits taken from MNIST\@.}
\label{tab:MNIST-results}
\medskip
\begin{tabular}{lc}
\toprule
{Model} & {Test LogL}\\
\midrule
MoBernoullis K=10 & $-168.95$ \\
MoBernoullis K=500 & $-137.64$ \\
RBM (500 h, 25 CD steps) approx. & $-86.34$\\
DBN 2hl approx. & $\mathbf{-84.55}$\\
NADE 1hl (fixed order) & $-88.86$ \\
NADE 1hl (fixed order, RLU, minibatch) & $-88.33$ \\
NADE 1hl (fixed order, sigm, minibatch) & $-88.35$  \\
\hline
NADE 1hl (no input masks)& $-99.37$ \\
NADE 2hl (no input masks)& $-95.33$ \\
NADE 1hl & $-92.17$ \\
NADE 2hl & $-89.17$ \\
NADE 3hl & $-89.38$ \\
NADE 4hl & $-89.60$ \\
EoNADE 1hl (2 orderings) & $-90.69$ \\
EoNADE 1hl (128 orderings) & $-87.71$ \\
EoNADE 2hl (2 orderings) & $-87.96$ \\
EoNADE 2hl (128 orderings) & $-85.10$ \\
\bottomrule
\end{tabular}
\end{center}
\end{table}

\begin{figure}
\centerline{\includegraphics[width=\columnwidth]{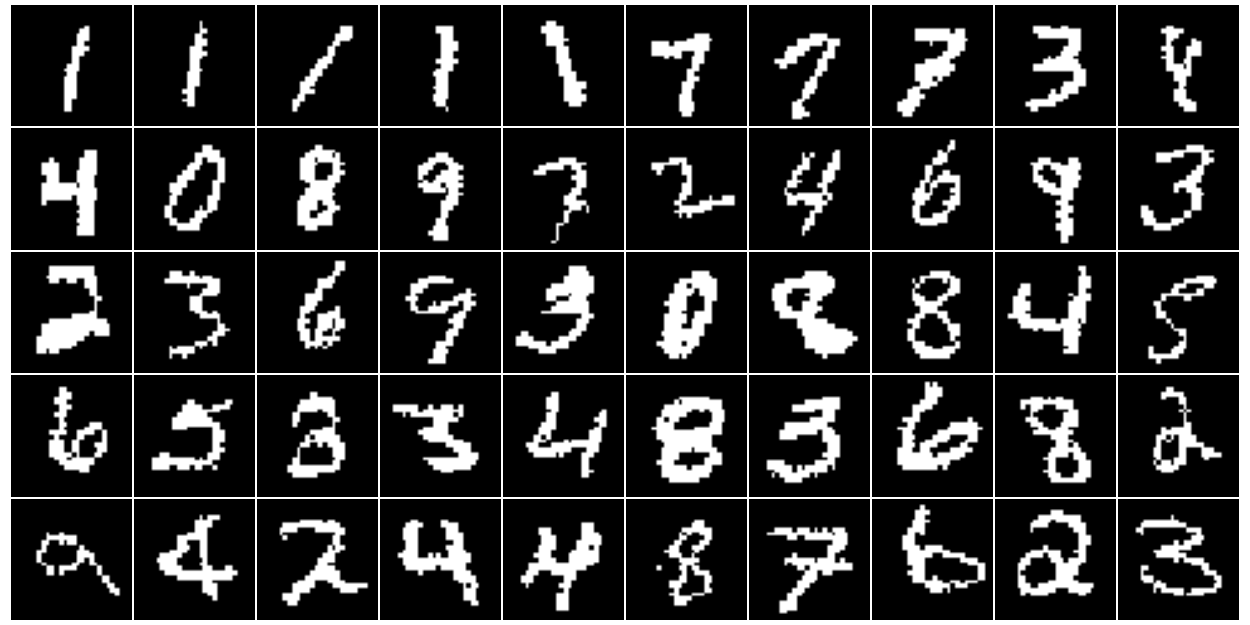}}
\centerline{\includegraphics[width=\columnwidth]{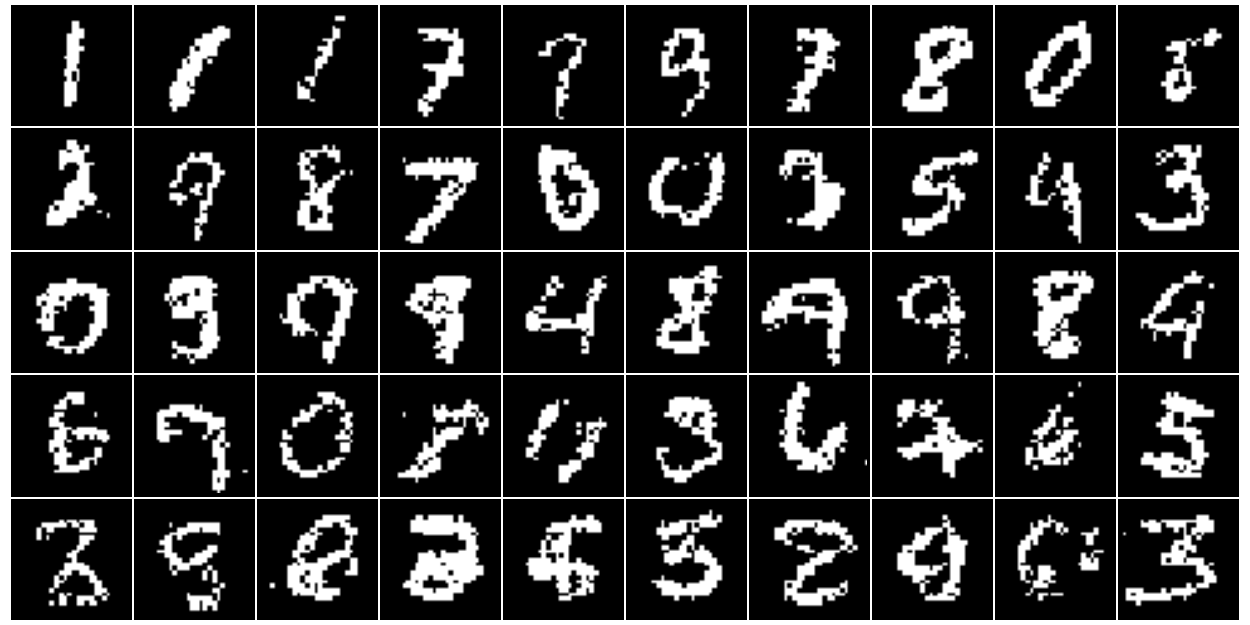}}
\caption{\textbf{Top:} 50 examples from binarized-MNIST ordered by decreasing
likelihood under a 2-hidden-layer NADE\@. \textbf{Bottom:}~ 50 samples from a
2-hidden-layer NADE\@, also ordered by decreasing likelihood under the model.}
\label{fig:MNIST-samples}
\end{figure}

\begin{figure}
\centerline{\includegraphics[width=0.8\columnwidth]{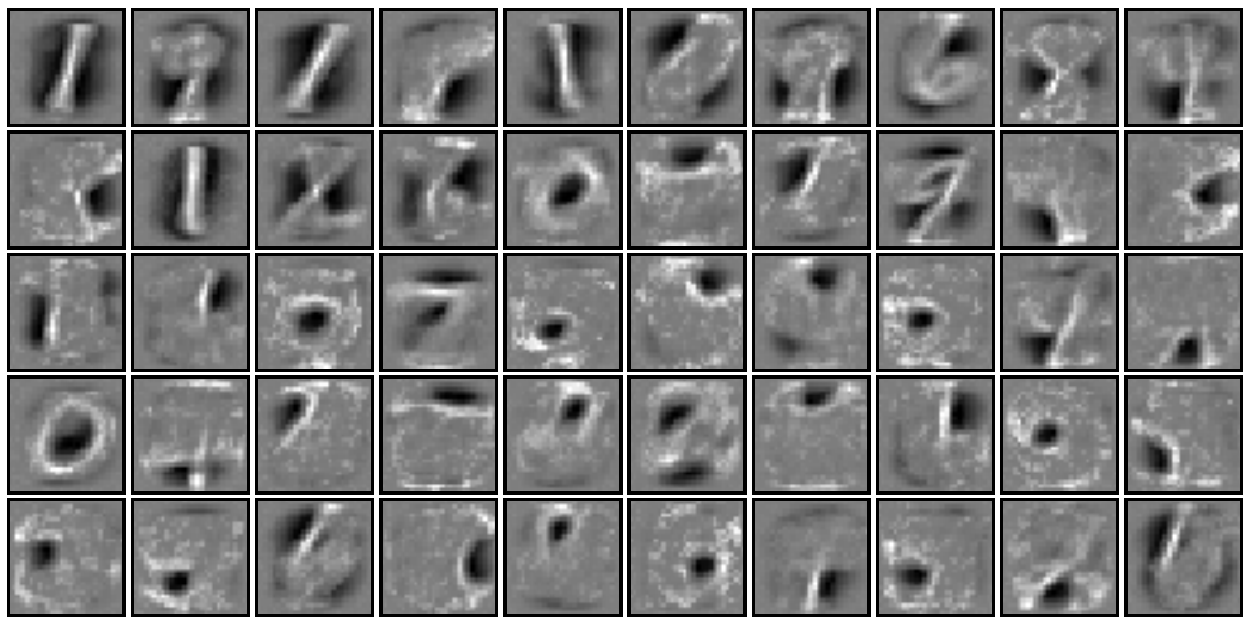}}
\centerline{\includegraphics[width=0.8\columnwidth]{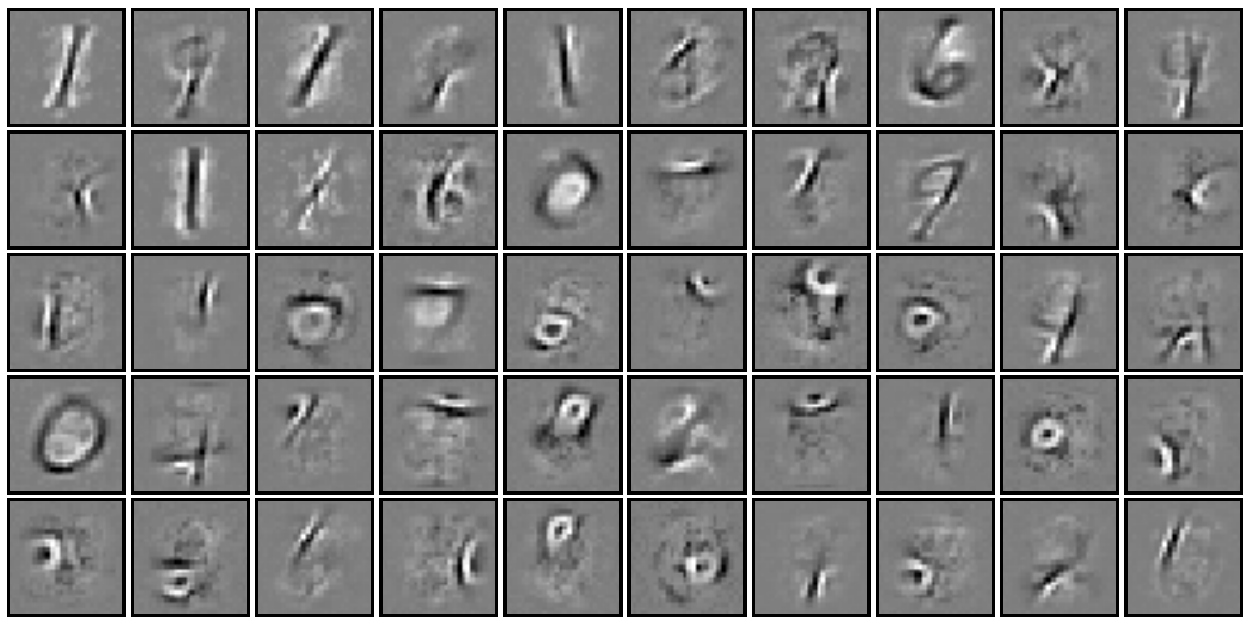}}
\caption{\textbf{Top:}50 receptive fields (columns of $\mat{W}$) with the
biggest L2 norm.
\textbf{Bottom:} Associated receptive fields to the input masks.}
\label{fig:MNIST-rfs}
\end{figure}

Samples from a 2 hidden layer (500 hidden units per layer) NADE trained using
the order-agnostic method are shown in Figure~\ref{fig:MNIST-samples}.
Most of the samples can be identified as digits.
Figure~\ref{fig:MNIST-rfs} shows some receptive fields from the model's first
hidden layer (\ie columns of $\mat{W}$).  Most of the receptive fields resemble
pen strokes. We also show their associated receptive fields on the input masks
. These can be thought of as biases that activate or
deactivate a hidden unit. Most of them will activate the unit when the input
mask contains a region of unknown values (zeros in the input mask) flanked by a
region of known values (ones in the input mask).

\begin{figure}
\centerline{\includegraphics[width=\columnwidth]{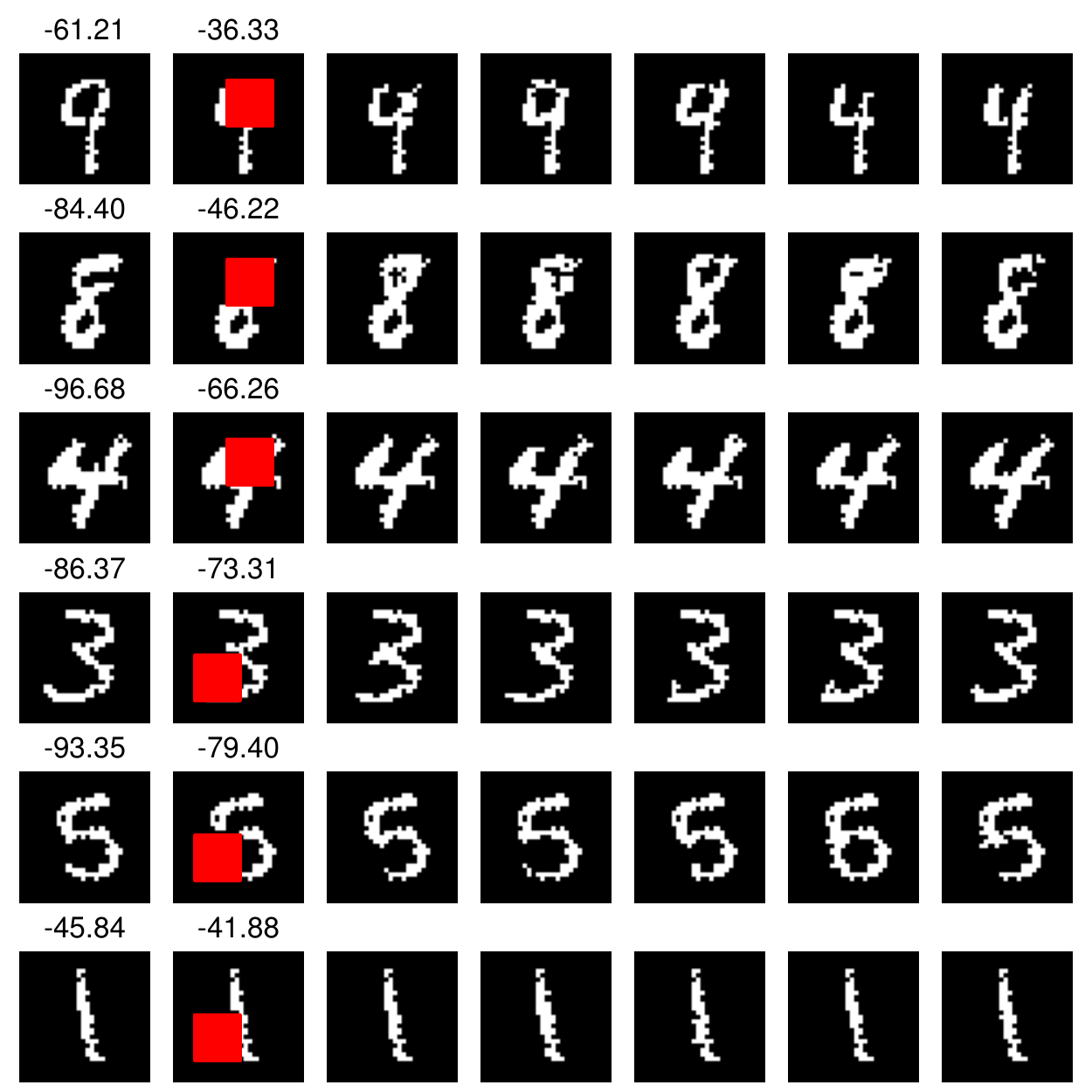}}
\caption{Example of marginalization and sampling. First column shows five
examples from the test set of the MNIST dataset. The second column shows
the density of these examples when a random 10 by 10 pixel region is
marginalized. The right-most five columns show samples for the hollowed
region. Both tasks can be done easily with a NADE where the pixels to marginalize are at the end of the ordering.}
\label{fig:mnist-inpainting}
\end{figure}

Having at our disposal a NADE for each possible ordering of the inputs makes it
easy to perform any inference task. In Figure~\ref{fig:mnist-inpainting} we show
examples of marginalization and imputation tasks. Arbitrarily chosen
regions of digits in the MNIST test-set are to be marginalized or sampled from.
An RBM or a DBN would require an exponential number of operations to
calculate either the marginal density or the density of the complete images.
A NADE trained on a fixed ordering of the variables would be able to easily
calculate the densities of the complete images, but would require approximate
inference to calculate the marginal densities. Both an RBM and a fixed-order
NADE require MCMC methods in order to sample the hollowed regions.
However, with our order-agnostic training procedure we can easily calculate the
marginal densities and sample the hollowed regions in constant time just by
constructing a NADE with a convenient ordering of the pixels.

\subsection{Real-valued datasets}
We also compared the performance of RNADEs trained with our order-agnostic
procedure to RNADEs trained for a fixed ordering. We start by comparing the
performance on three low-dimensional UCI datasets~\cite{Bache+Lichman:2013} of heterogeneous
data, namely: \emph{red wine}, \emph{white wine} and \emph{parkinsons}. We dropped the
other two datasets tested by \citet{UriaB2013}, because some of their dimensions only
take a finite number of values even if those are real-valued.
We report the test-log-likelihood on 10 folds of the dataset, each with 90\%
of the data used for training and 10\% for testing. All experiments use
normalized data. Each dimension is normalized separately by subtracting its
training-set average and dividing by its standard deviation.

Experimental details follow. Learning rate and weight decay rates were chosen
by per-fold cross-validation; using grid search. One ninth of the training set
examples were used for validation purposes. Once the hyperparameter values had
been chosen, a final experiment was run using all the training data. In order to
prevent overfitting, training was stopped when observing a training likelihood higher than
the one obtained at the optimal stopping point in the corresponding validation
run. All RNADEs trained had a mixture of 20 Gaussian components for output, and
were trained by stochastic gradient descent on $\loss$. We fixed the number of
hidden units to 50, following \citet{UriaB2013}. The learning rate was chosen among
$\{0.02, 0.005, 0.002, 0.0005\}$ and the weight decay rate among $\{0.02, 0.002,
0\}$.

The results are shown in Table~\ref{tab:RNADE-UCI-results}. RNADEs trained using our
procedure obtain results close to those of RNADEs trained for a fixed ordering
on the \emph{red wine} and \emph{white wine} datasets. On the Parkinsons dataset,
RNADEs trained for a fixed ordering perform better. Ensembles of
RNADEs obtained better statistical performance on the three datasets.

\begin{table*}
\vspace*{-0.1in}
\begin{center}
\caption{Average test log-likelihood for different models on three real-valued
UCI datasets. Baselines are taken from~\cite{UriaB2013}.}
\label{tab:RNADE-UCI-results}
\medskip
\begin{tabular}{lcccccccc}
\toprule
{Model} & {Red wine} & {White wine} & {Parkinsons} \\
\midrule
{Gaussian} & $-13.18$ & $-13.20$ & $-10.85$ \\
{MFA} & $-10.19$ & $-10.73$ & $-1.99$ \\
{RNADE (fixed)} & $\mathbf{-9.36}$ & $\mathbf{-10.23}$ & $\mathbf{-0.90}$ \\
\hline
{RNADE 1hl}
&$-9.49$
&$-10.35$
&$-2.67$
\\
{RNADE 2hl}
&$-9.63$
&$\mathbf{-10.23}$
&$-2.19$
\\
{RNADE 3hl}
&$-9.54$
&$\mathbf{-10.21}$
&$-2.13$
\\
{RNADE 1hl 2 ord.}
&$-9.07$
&$-10.03$
&$-1.97$
\\
{RNADE 2hl 2 ord.}
&$-9.13$
&$-9.84$
&$-1.42$
\\
{RNADE 3hl 2 ord.}
&$-8.93$
&$-9.79$
&$-1.39$
\\
{RNADE 1hl 16 ord.}
&$-8.95$
&$-9.94$
&$-1.73$
\\
{RNADE 2hl 16 ord.}
&$-8.98$
&$-9.69$
&$-1.16$
\\
{RNADE 3hl 16 ord.}
&$\mathbf{-8.76}$
&$\mathbf{-9.67}$
&$-1.13$
\\
\bottomrule
\end{tabular}
\end{center}
\end{table*}

We also measured the performance of our new training procedure on 8 by 8 patches
of natural images in the BSDS300 dataset. We compare the performance of RNADEs
with different number of hidden layers trained with our procedure against a
one-hidden layer RNADE trained for a fixed ordering~\cite{UriaB2013}, and with
mixtures of Gaussians, which remain the state of the art in this
problem~\cite{Zoran2012}.

We adopted the setup described by \citet{UriaB2013}. The average intensity of
each patch was subtracted from each pixel's value. After this, all datapoints lay
on a 63-dimensional subspace, for this reason only 63 pixels were modelled,
discarding the bottom-right pixel.

Experimental details follow. The dataset's 200 training image set was
partitioned into a training set and a validation set of 180 and 20 images
respectively. Hyperparameters were chosen by preliminary manual search on the
model likelihood for the validation dataset. We used a mixture of 10
Gaussian components for the output distribution of each pixel. All hidden layers
were fixed to a size of 1000 units. The minibatch size was set to 1000. Training was run
for 2000 iterations, each consisting of 1000 weight updates. The initial learning
rate was set to 0.001. Pretraining of hidden layers was done for 50 iterations.

The results are shown in Table~\ref{tab:BSDS-results}. RNADEs with less than 3
hidden layers trained using our order-agnostic procedure obtained lower statistical
performance than a fixed-ordering NADE and a mixture of Gaussians. However RNADEs
with more than 3 layers are able to beat both baselines and obtain what are, to
the extent of our knowledge, the best results ever reported on this task.
Ensembles of RNADEs also show an improvement in statistical performance compared
to the use of single RNADEs.

\begin{figure}
\vspace*{-0.1in}
\centerline{\includegraphics[width=0.75\columnwidth]{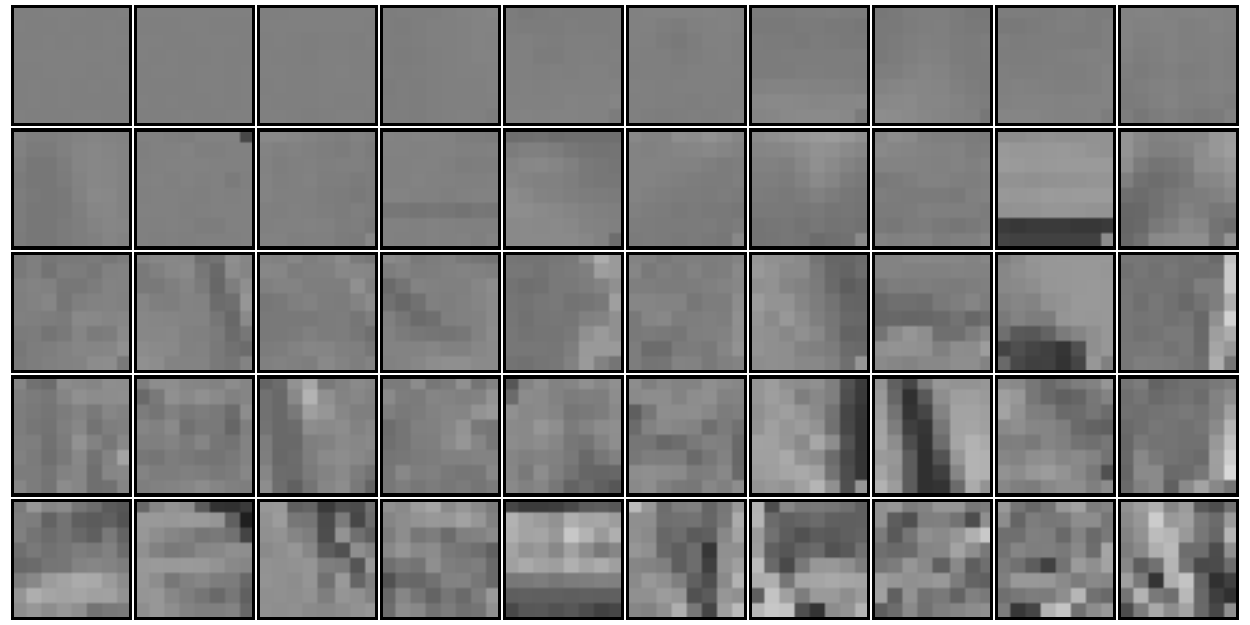}}
\centerline{\includegraphics[width=0.75\columnwidth]{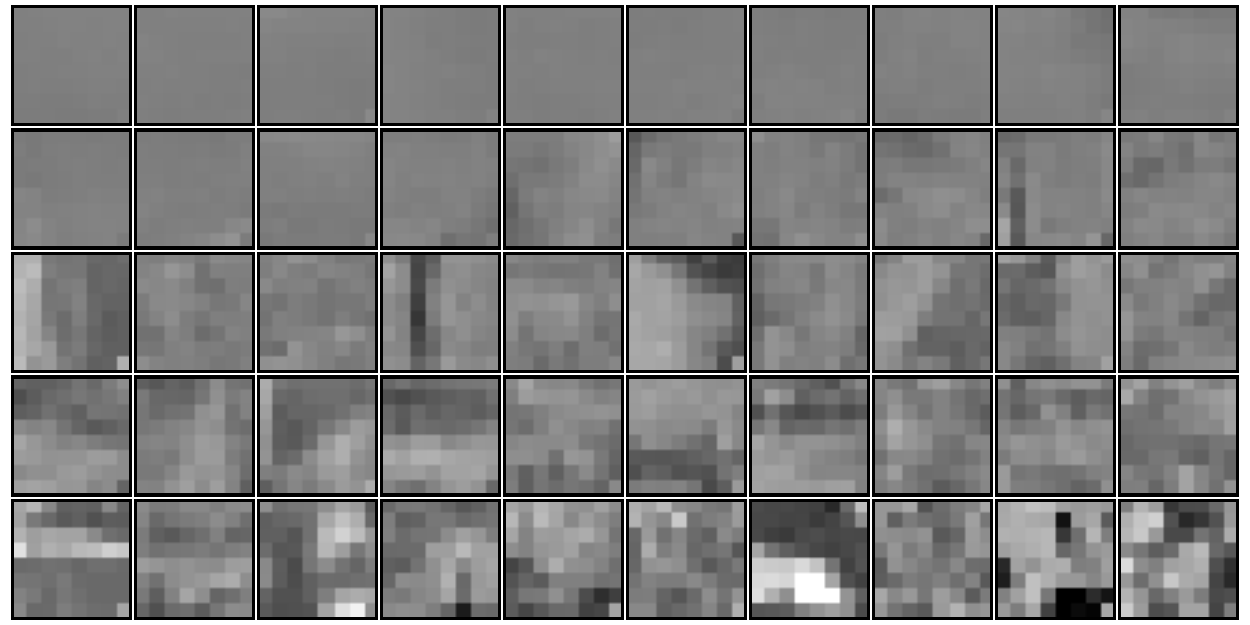}}
\caption{\textbf{Top:} 50 examples of $8\!\times\!8$ patches in the BSDS300
dataset ordered by decreasing likelihood under a 6-hidden-layer NADE\@. \textbf{Bottom:}~
50 samples from a 6-hidden-layer NADE\@.%
}
\label{fig:natural-image-samples}
\vspace*{-0.3in}
\end{figure}

No signs of overfitting were observed. Even when using 6 hidden layers, the cost
on the validation dataset never started increasing steadily during training.
Therefore it may be possible to obtain even better results using more hidden layers or
more hidden units per layer.
\begin{table}
\vspace*{-0.2in}
\begin{center}
\caption{Average test-set log-likelihood for several models
trained on 8 by 8 pixel patches of natural images taken from the BSDS300
dataset. Note that because these are log probability densities they are positive,
higher is better.}
\label{tab:BSDS-results}
\medskip
\begin{tabular}{lc}
\toprule
{Model} & {Test LogL}\\
\midrule
MoG $K\!=\!200$\ \cite{Zoran2012} & $152.8$ \\
RNADE 1hl (fixed order) & $152.1$ \\
\hline
RNADE 1hl & $143.2$ \\
RNADE 2hl & $149.2$ \\
RNADE 3hl & $152.0$ \\
RNADE 4hl & $153.6$ \\
RNADE 5hl & $154.7$\\
RNADE 6hl & $\mathbf{155.2}$ \\
EoRNADE 6hl 2 ord.  & $156.0$ \\
EoRNADE 6hl 32 ord. & $\mathbf{157.0}$ \\
\bottomrule
\end{tabular}
\end{center}
\vspace*{-0.3in}
\end{table}
Samples from the 6 hidden layers NADE trained in an order-agnostic manner are
shown in Figure~\ref{fig:natural-image-samples}.

\section{Conclusions}
We have introduced a new training procedure that simultaneously fits a NADE for
each possible ordering of the dimensions. In addition, this new training
procedure is able to train deep versions of NADE with a linear increase
in computation, and construct ensembles of NADEs on the fly without incurring
any extra training computational cost.

NADEs trained with our procedure
outperform mixture models in all datasets we have investigated. However, for
most datasets several hidden layers are required to surpass or equal
the performance of NADEs trained for a fixed ordering of the variables.
Nonetheless, our method allows fast and exact marginalization and
sampling, unlike the rest of the methods compared.

Models trained using our order-agnostic procedure obtained what are, to the best
of our knowledge, the best statistical performances ever reported on the
\emph{BSDS300} 8$\times$8-image-patches datasets. The use of ensembles of NADEs,
which we can obtain at no extra training cost and have a mild effect on
test-time cost, improved statistical performance on most datasets analyzed.
\vspace{-0.5ex}

\subsubsection*{Acknowledgments}\vspace{-0.5ex}
\begin{small}We thank John Bridle and Steve Renals for useful
discussions.\end{small}

\FloatBarrier
\bibliography{main}
\bibliographystyle{icml2014}

\end{document}